# Matching Writers to Content Writing Tasks

Narayana Darapaneni[1], Chandrashekhar Bhakuni[2], Ujjval Bhatt[3], Khamir Purohit[4], Vikas Sardna[5], Prabir Chakraborty[6], and Anwesh Reddy Paduri[7]

[1] Northwestern University/Great Learning, Evanston, US
[2-7]Great Learning, Bangalore, India

anwesh@greatlearning.in

**Abstract.** Businesses need content. In various forms and formats and for varied purposes. In fact, the content marketing industry is set to be worth $412.88 billion by the end of 2021. However, according to the Content Marketing Institute, creating engaging content is the #1 challenge that marketers face today. We understand that producing great content requires great writers who understand the business and can weave their message into reader (and search engine) friendly content. In this project, the team has attempted to bridge the gap between writers and projects by using AI and ML tools. We used NLP techniques to analyze thousands of publicly available business articles (corpora) to extract various defining factors for each writing sample. Through this project we aim to automate the highly time-consuming, and often biased task of manually shortlisting the most suitable writer for a given content writing requirement. We believe that a tool like this will have far reaching positive implications for both parties - businesses looking for suitable talent for niche writing jobs as well as experienced writers and Subject Matter Experts (SMEs) wanting to lend their services to content marketing projects. The business gets the content they need, the content writer/ SME gets a chance to leverage his or her talent, while the reader gets authentic content that adds real value.

## 1 Introduction

We use data-driven techniques to identify some of the many aspects that make a writer unique[1]. We then use concepts like Myers–Briggs Type Indicator (MBTI) to extract the traits and abilities of an author, using his/ her write-up.

In order to analyze the writing styles, we started exploring publicly available corpora of long-form content formats like blogs and articles. Based on our subject matter understanding and data scraping feasibility, we zero-ed in on websites like Harvard Business School (https://hbswk.hbs.edu/) and Entrepreneur India (http://entrepreneur.com/)

that publish articles in sub-domains like Management, Finance, Strategy, and Leadership.

We built our own dataset using the Beautiful Soup package of Python[2]. Using this method, we were able to parse approximately 40,000 articles. Data points such as page URL, article headline, article text, business domain, name of the author were the ones of interest for us.

Textual data, like long-form articles, requires a significant amount of cleaning and pre-process to retain the relevant text[4]. The steps also vary as per the content source. For this project, we cleaned out several unimportant texts. In addition, extra focus was required to identify, and weed-out special characters and junk elements introduced due to various encoding techniques (like UTF-8, UTF-16 and UTF-32).

During the project, we encountered, explored and implemented various algorithms, methods, and thought processes[8]. The nature of our goal warranted that we work with multiple models in parallel; and eventually combine them to offer a prediction.

We started with multiple NLP libraries and simple ML models like TF-IDF encoder, MultinomialNB classifier, and went on to make complex and multi-layered models that employ Deep Learning, Neural Networks, Universal Sentence Encoder, LSTM[7] with GloVe embeddings, fastText, and its many variations.

In all, the project was a journey of discovery and learning where we developed and experimented with multiple models - either to enhance relevance, or to improve accuracy.

## 2 Step-by-step Walkthrough of the Solution

**Fig. 1.** Project Execution Workflow

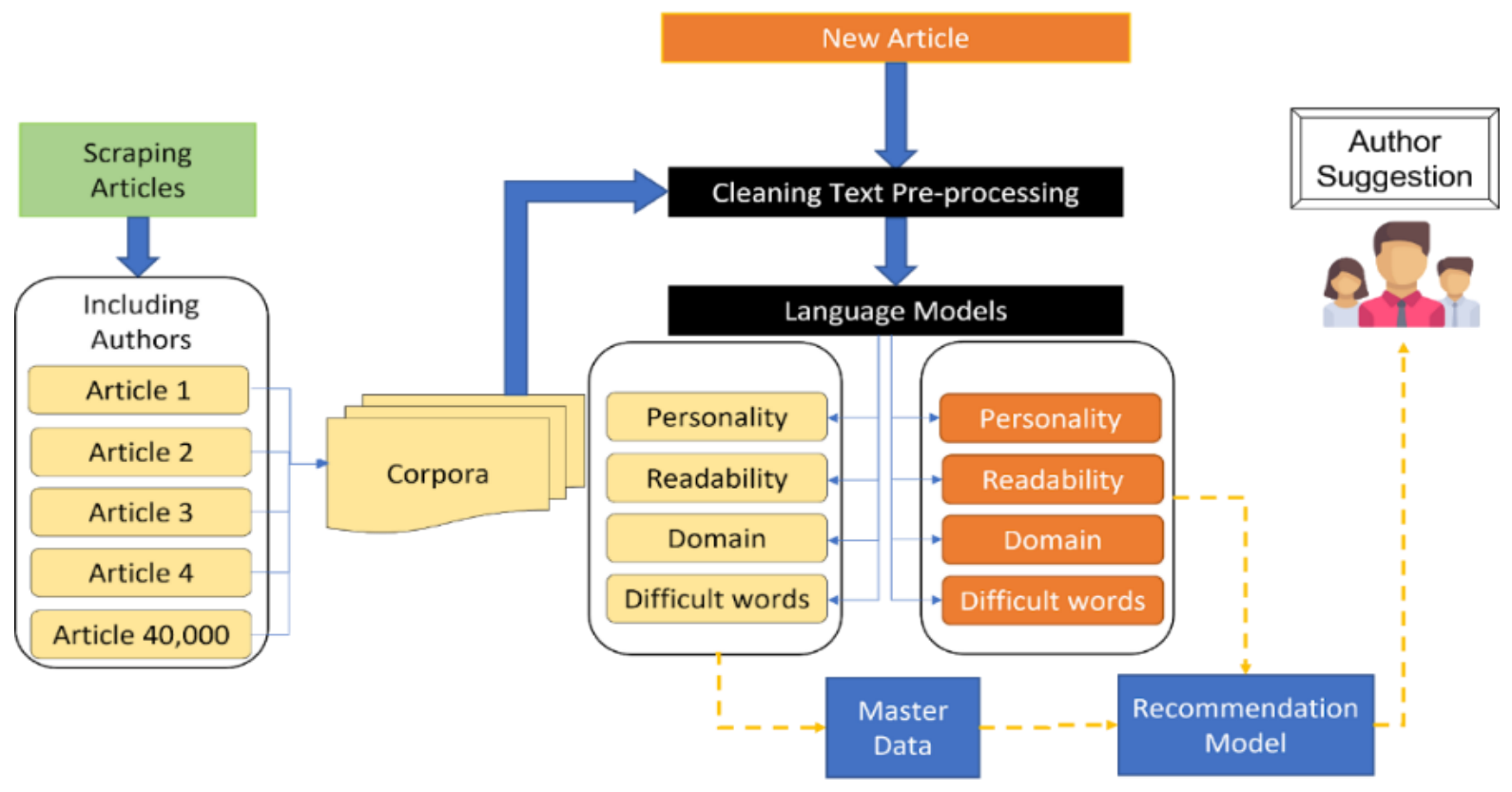

# 3 Data Collation and Cleaning

Include all relevant visualizations that support the ideas/ insights that you gleaned from the data. Scraped data from websites were cleaned up.

| Article URL | Headline | Article Text | Sub Domain | Domain | Author |
|---|---|---|---|---|---|
| https://www.entrepreneur.com/article/370874 | \n\t\t\t\tIs Launching a New Brand the Right M... | \n\n\n\t\t\t\t\t\t\t\tJune\n\t\t\t\t\t\t\t\t10... | Branding | Marketing | Melissa Packham |
| https://www.entrepreneur.com/article/372254 | \n\t\t\t\t3 Reasons Simple Isn't Always Better... | \n\n\n\t\t\t\t\t\t\t\tJune\n\t\t\t\t\t\t\t\t10... | Branding | Marketing | Zaheer Dodhia |
| https://www.entrepreneur.com/article/368356 | \n\t\t\t\tFrom Idea to Revenue: A Six-Step For... | \n\n\n\t\t\t\t\t\t\t\tJune\n\t\t\t\t\t\t\t\t10... | Launching a Business | Marketing | Jessica O'Connell |
| https://www.entrepreneur.com/article/373825 | \n\t\t\t\tAvoid This Common Mistake When Writi... | \n\n\n\t\t\t\t\t\t\t\tJune\n\t\t\t\t\t\t\t\t10... | Writing a Book | Marketing | R. Paulo Delgado |
| https://www.entrepreneur.com/article/372679 | \n\t\t\t\tIs it Time to Say RIP to RFPs? \n\t\t\t | \n\n\n\t\t\t\t\t\t\t\tJune\n\t\t\t\t\t\t\t\t9,... | Entrepreneurs | Marketing | Heather Ripley |

**Fig. 2.** Scrapped Data

| | Domain | Cleaned_Article | Cleaned_Headline | Cleaned_Author | Word_Count | Article_Size |
|---|---|---|---|---|---|---|
| 0 | Marketing | One of the most common challenges I see client... | Is Launching a New Brand the Right Move for Yo... | Melissa Packham | 1013 | Large |
| 1 | Marketing | Lets say that, like me, youve been thinking ab... | 3 Reasons Simple Isn't Always Better When It C... | Zaheer Dodhia | 882 | Medium |
| 2 | Marketing | If youre like most entrepreneurs, you may have... | From Idea to Revenue A Six-Step Formula to Lau... | Jessica O'Connell | 1409 | Large |
| 3 | Marketing | I had an interesting project come across my de... | Avoid This Common Mistake When Writing Your Fi... | R. Paulo Delgado | 952 | Medium |
| 4 | Marketing | In a world where personal touch matters, why a... | Is it Time to Say RIP to RFPs | Heather Ripley | 723 | Medium |

**Fig. 3.** Cleaned and Headline Article Text, with calculated Article Size, etc.

We have 29,631 articles the article distribution is:

**Table 1.** Data distribution

**Size-Wise**

| | |
|---|---|
| **Small (<500)** | 6164 |
| **Medium (501 - 1000)** | 16576 |
| **Large (>1000)** | 6891 |

**Domain-Wise**

| | |
|---|---|
| **Finance** | 4600 |
| **Leadership** | 6267 |
| **Marketing** | 7467 |
| **Strategy** | 5205 |
| **Technology** | 6092 |

**Cross Distribution**

| | **Large** | **Medium** | **Small** |
|---|---|---|---|
| **Finance** | 1031 | 2553 | 1016 |
| **Leadership** | 1736 | 3889 | 642 |
| **Marketing** | 1632 | 4572 | 1263 |
| **Strategy** | 1578 | 3007 | 620 |
| **Technology** | 914 | 2555 | 2623 |

We have the highest number of articles for Marketing, and the least number of articles for Finance. We also observe that most articles belong to the Medium-size category.

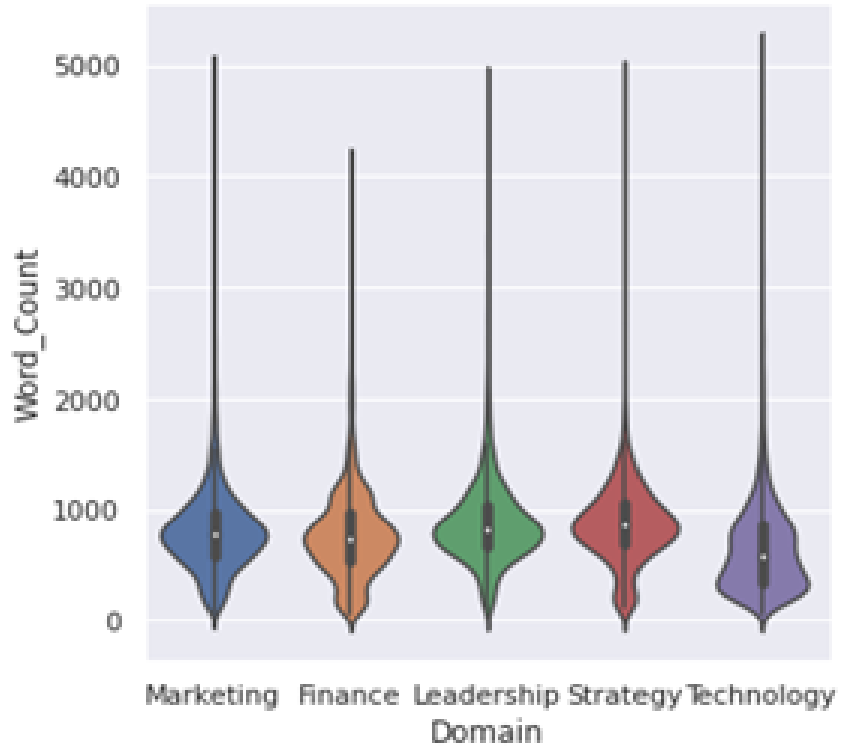

**Fig. 4.** Domain wise Word-count distribution

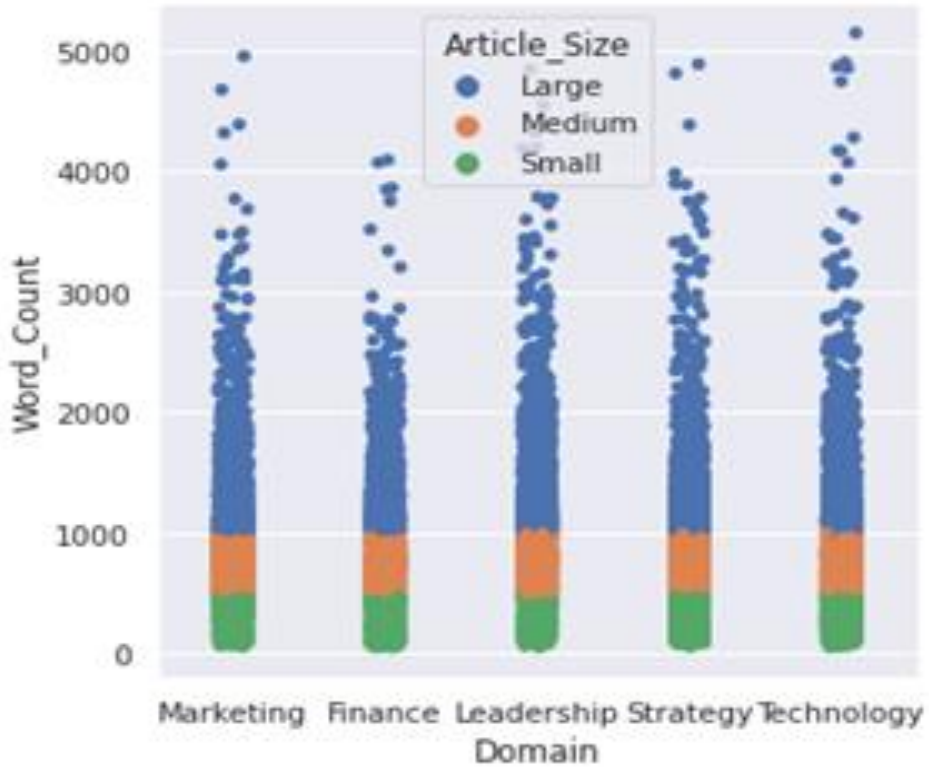

**Fig. 5.** Domain and Article size wise Word-count distribution

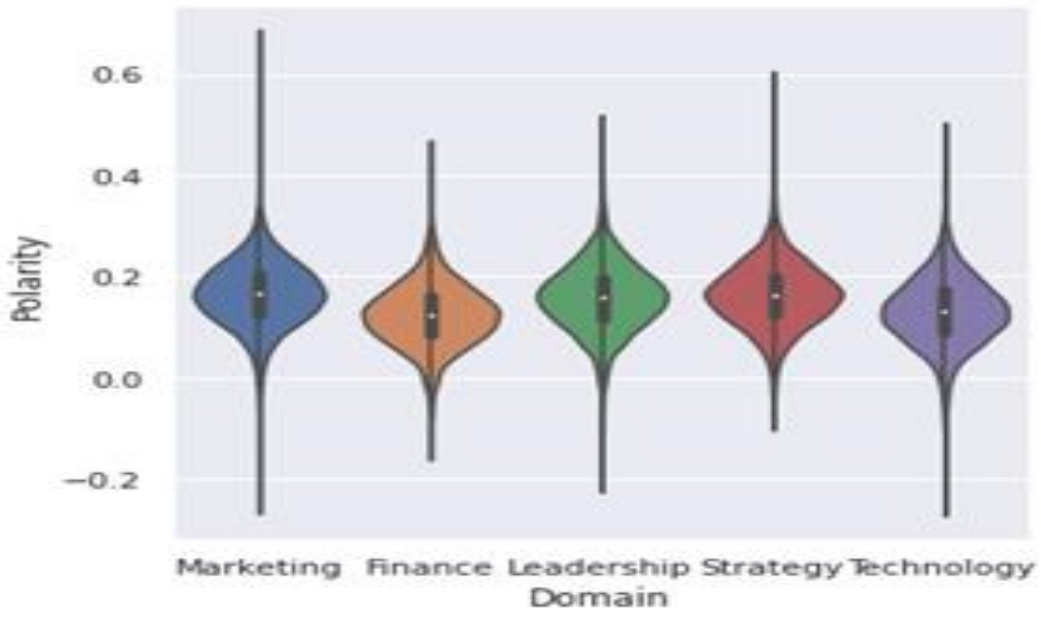

**Fig. 6.** Domain wise Polarity distribution

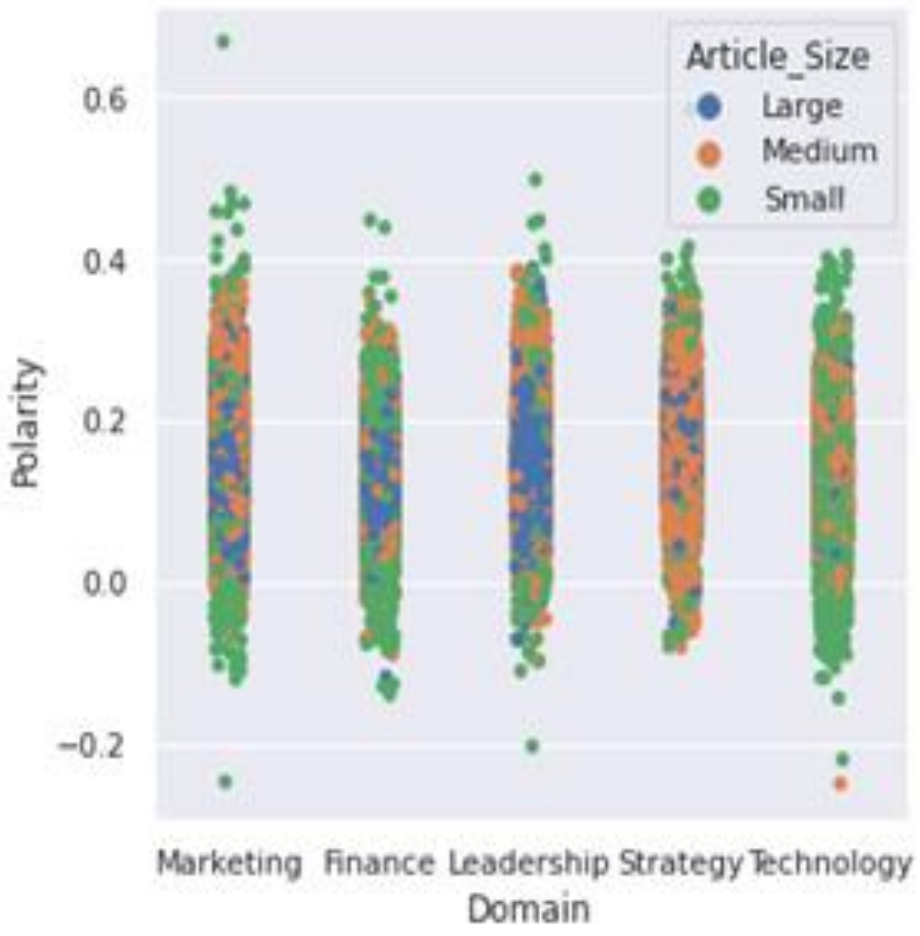

**Fig. 7.** Domain and Article size wise Polarity distribution

**Table 2.** Sentiment-wise Distribution

| | | Domain | | | | |
|---|---|---|---|---|---|---|
| | | **Finance** | **Leadership** | **Marketing** | **Strategy** | **Technology** |
| **Sentiment** | **Negative** | 126 | 77 | 85 | 44 | 158 |
| | **Neutral** | 1470 | 1013 | 934 | 698 | 1740 |
| | **Positive** | 3004 | 5177 | 6448 | 4463 | 4194 |

After the sentiment analysis we observe that Positive articles have the highest percentage followed by Neutral and Negative.

# 4 Build Up To The Final Solution

**Table 3.** Steps Taken

| **Steps Taken to Solve the Problem** | **Findings at Each Stage** | **Implications for Next Steps** |
|---|---|---|
| Scraping data from two sources: HBS and Entrepreneur.com for five domains. Collected approx. 40,000 articles | Different sources had various formats of the articles and different places for author names, etc.<br>Scraping the data using Beautiful Soup library, for different source websites | Cleaned out items like static headers and footers, publisher information, list of references, and web page elements like CTA buttons, banners, hyperlink texts, special characters and junk elements. |
| | Scraping led to duplicate articles with garbled text | |
| Cleaning the data using Beautiful Soup and Regular Expressions (Regex) by removing multiple special characters, duplicate records, articles with missing authors or small length (< 50 words), etc. | Dropping articles with duplicate values and missing values.<br>Cleaning of text data required multiple iterations. | Total articles came down to approx. 29,000.<br>With major reduction in the Finance domain. |
| More than 15 features for articles were calculated using NLP libraries NLTK, TextSTAT and TextBlob. | Flesch_Reading_Ease was chosen for computing the readability[9].<br>Difficult words with more than two syllables were calculated and an appropriate feature class was created. | The key features were identified for Author recommendation. |
| To derive MBTI for each article, data (essay corpus) from University of Antwerp was analyzed. | We have used the same MBTI data to train our model on various algorithms. Out of which LSTM gave the best results. | We then used the model to predict the personality traits of the authors in our dataset. |
| To predict the domain class, we developed multiple models, starting with simple Naive Bayes ML model[5], to Transfer Learning models, USE, Conv1D + bi directional LSTM, character embedding[10], hybrid model with sentence and character embedding and GloVe embedding with LSTM. | Embedding only covered partial vocabulary. For example, GloVe only offered 40% of vocabulary of the articles in our dataset.<br>The models were giving lower accuracy and F1 score specifically for Strategy and Leadership domains. | New words and acronyms such as “GDP”, ‘Facebook”, “Tesla”, “Amazon”, etc., were not part of any embeddings.<br>Recommendation for authors for Leadership and Strategy might not be accurate. |
| Steps Taken to Solve the Problem | Findings at Each Stage | Implications for Next Steps |
| Combining all the features and making master data for feeding into our recommendation system. | Out of 18+ parameters, we identified the parameters that would be considered for author recommendations. | Domain, Readability, MBTI, and Density of Difficult Words, were the four features selected. |
| Recommendations using KNN and Cosine Similarity[11] for the selected features by converting the data into One Hot Encoding. | KNN could give effective predictions since the data was sparse.<br>Cosine Similarity was calculated, and content-based recommendations were derived. | Master data with all the authors was created. It was subsequently used for evaluating the new article. |
| Features for a new article were derived with (a) Saved models and (b) NLP library. The features were One Hot Encoded and fed to the recommendation system. | We could generate top "n" recommended authors for any new requirement. | |

# 5 Model Evaluation

As described above, the final model consists of several pieces that help generate the features for existing articles and any new article sample received. Based on these parameters a recommendation system will predict the best match from the existing authors.

- Domain Classification (NLP Model)
- Writer Personality (MBTI) Classification (NLP Model)
- Rule-based Computation
- Readability Score: Using Flesch_Reading_Ease score
- Difficult-Word Density

We will describe the models one-by-one:

## 5.1 Domain Classification

The below mentioned algorithms have been used/ explored for finding various aspects of the text articles:

1. Multinomial Naive Bayes (ML Baseline Model) with TF-IDF [6]
2. LSTM with GloVe Embedding [12]
3. Conv1D with Tensorflow Token Embedding [17], [19], [20]

4. Deep Neural Network with Pretrained Token Embedding using Tensorflow Hub's Universal Sentence Encoder (USE) [5]
5. Conv1D with Character Embedding [18], [13], [14]
6. Bidirectional LSTM with Hybrid Embedding - Pretrained Token Embedding (USE) + Character Embedding
7. GRU [15], [16]
8. BERT/ DistilBERT Tokenizers [12]
9. HuggingFace Transformers

The model was trained on existing data and is able to predict the Domain for a new article. We experimented with multiple embedding techniques, Transfer Learning from pre-trained embeddings and various classification models. We started with a simple Naive Bayes classifier and moved to complex deep neural networks, Bi directional LSTM and Conv1D. We finally picked the best performing model for predicting the domain of a new article.

Accuracy, Precision, Recall and F1-score of various experimental models on test data were tabulated and the model with highest F1-score was picked as the best model.

**Table 4.** Table

| Model | Model Description | Accuracy % | F1-Score |
|---|---|---|---|
| **Baseline: TF-IDF + MultinomialNB** | Text vectorization by TF-IDF followed by Naive Bayes Classifier | 61.77 | 0.55 |
| **Model 1: Token Embed + Conv1D** | Text vectorization followed by word embedding through Keras fed to Conv1D, Global Max and Global Average Pooling layers fed to output dense layer:<br>• Total params: 3,881,669<br>• Trainable params: 3,881,669<br>• Non-trainable params: 0 | 64.44 | 0.65 |
| **Model 2: Pre-trained USE Embedding + Dense** | Text vectorization followed by pre-trained Embeddings from Universal Sentence Encoder (USE) fed to the output layer.<br>• Total params: 256,864,133<br>• Trainable params: 66,309<br>• Non-trainable params: 256,797,824 | 73.82 | 0.73 |
| **Model 3: Char Embed + Conv1D** | Text vectorization followed by word embedding through Keras fed to Conv1D layer and Global Max fed to output layer. | 20.85 | 0.07 |
| | • Total params: 10,139<br>• Trainable params: 10,139<br>• Non-trainable params: 0 | | |
| **Model 4:** | Combination of Character Embedding and USE Embedding fed into BiLSTM and output dense layer.<br>• Total params: 256,912,243<br>• Trainable params: 114,419<br>• Non-trainable params: 256,797,824 | 73.82 | 0.72 |
| **Model 5: GloVe Embed + BiLSTM + Time-distributed** | Text vectorization followed by word embedding through GloVe Embedding fed to BiLSTM and Time-distributed and output dense layer.<br>• Total params: 76,854,457<br>• Trainable params: 76,854,457<br>• Non-trainable params: 0 | 69.74 | 0.69 |
| **fastText** | fastText model with trigrams and one-vs-all loss was used. | | 0.72 |

Both Model 2 and Model 4 gave very close output. However, Model 2 had a slightly higher F1-score and is simpler with only 66K trainable parameters compared to Model 4, which had 114K trainable parameters.

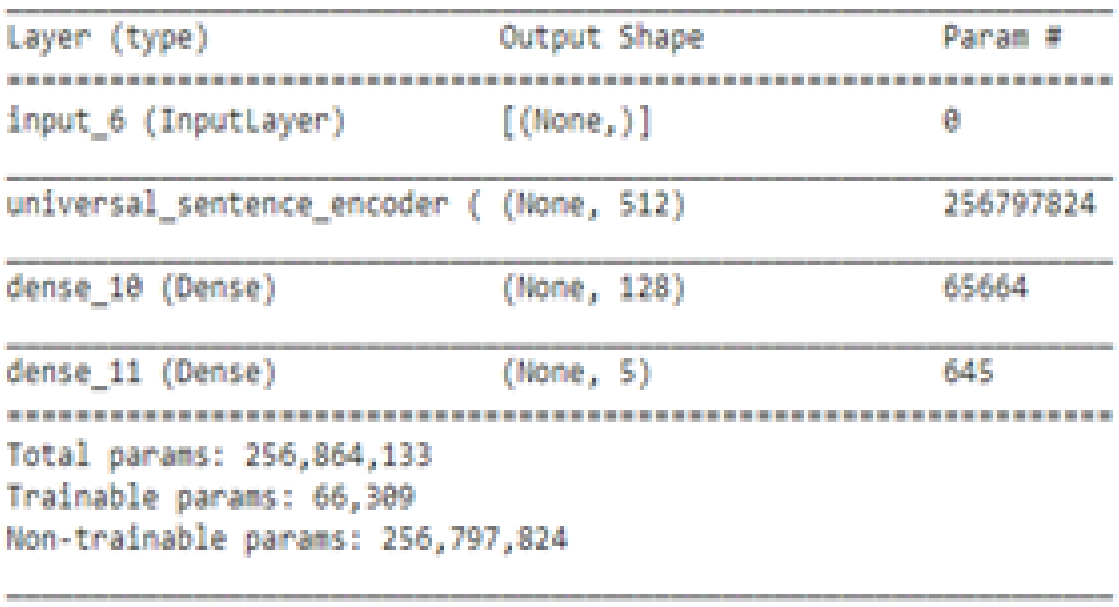

| Layer (type) | Output Shape | Param # |
|---|---|---|
| input_6 (InputLayer) | [(None,)] | 0 |
| universal_sentence_encoder ( | (None, 512) | 256797824 |
| dense_10 (Dense) | (None, 128) | 65664 |
| dense_11 (Dense) | (None, 5) | 645 |

Total params: 256,864,133
Trainable params: 66,309
Non-trainable params: 256,797,824

**Fig. 8.** Model 2 - Pretrained USE with Dense Layers

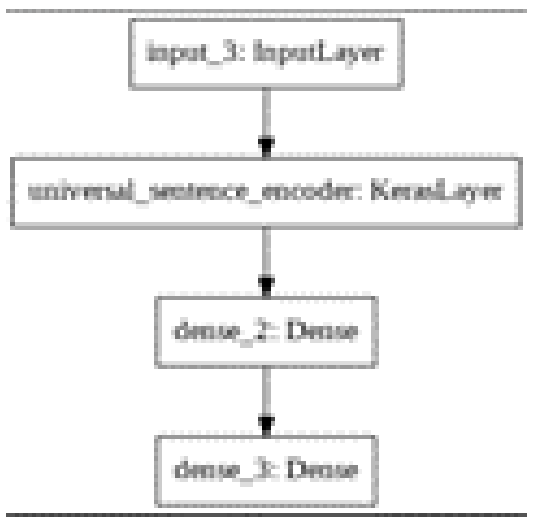

**Fig. 9.** Model 2 – flow diagram

| Layer (type) | Output Shape | Param # | Connected to |
|---|---|---|---|
| char_input (InputLayer) | [(None, 1)] | 0 | |
| token_input (InputLayer) | [(None,)] | 0 | |
| char_vectorizer (TextVectorizat | (None, 8914) | 0 | char_input[0][0] |
| universal_sentence_encoder (Ker | (None, 512) | 256797824 | token_input[0][0] |
| char_embed (Embedding) | (None, 8914, 25) | 1750 | char_vectorizer[2][0] |
| dense_18 (Dense) | (None, 128) | 65664 | universal_sentence_encoder[4][0] |
| bidirectional_1 (Bidirectional) | (None, 50) | 10200 | char_embed[2][0] |
| token_char_hybrid (Concatenate) | (None, 178) | 0 | dense_18[0][0]<br>bidirectional_1[0][0] |
| dropout_3 (Dropout) | (None, 178) | 0 | token_char_hybrid[0][0] |
| dense_19 (Dense) | (None, 200) | 35800 | dropout_3[0][0] |
| dropout_4 (Dropout) | (None, 200) | 0 | dense_19[0][0] |
| dense_20 (Dense) | (None, 5) | 1005 | dropout_4[0][0] |

Total params: 256,912,243
Trainable params: 114,419
Non-trainable params: 256,797,824

**Fig. 10.** Model 4 - Pre-trained USE Embedding with Char Embed and Bidirectional LSTM

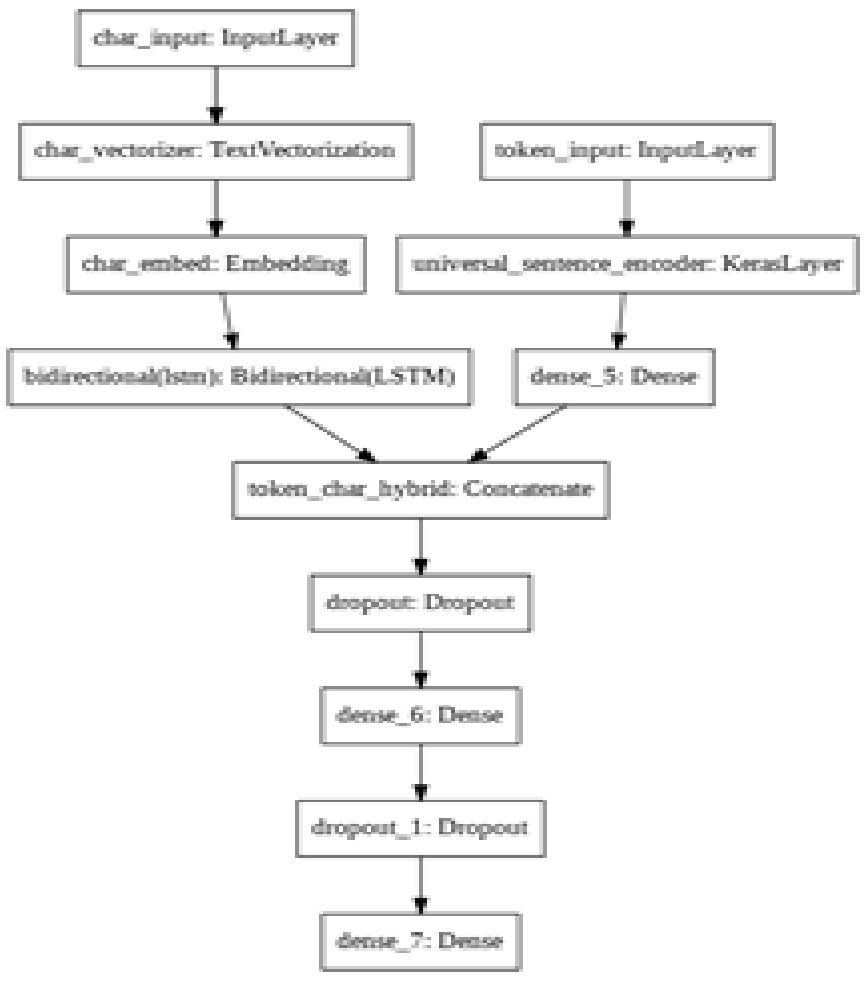

**Fig. 11.** Model 2 – flow diagram

Both these models use 256M+ non-trainable parameters from TFHub's Universal Sentence Encoder. We do not expect much variation and either of these two models can be used for prediction without much deviation in accuracy (74%). There is a chance that some new articles might be classified incorrectly.

### 5.2 Writer Personality using MBTI

The objective of MBTI [20-30] classification was to predict the personality of the writer based on the article the writer has written. We have trained our model using the MBTI data consisting of 8675 rows, wherein each article is tagged by its personality trait.

We have deployed different models using algorithms like Naive Bayes, Random Forest, XGBoost, Stochastic Gradient Descent, Logistic Regression, KNN, SVM and finally LSTM. Out of all the models we got the best accuracy (85%) using LSTM.

### 5.3 Readability

Flesch Reading-Ease Scores (FRES) were considered for evaluating the readability of the articles. The formula for FRES is:

**206.835-1.015 (total words/ total sentences) – 84.6 (total syllables/ total words) (1)**

Result is a number that corresponds with a U.S. grade level (Grade 5-10, College, Grad College, and Professional), which denotes an increasing level of complexity.

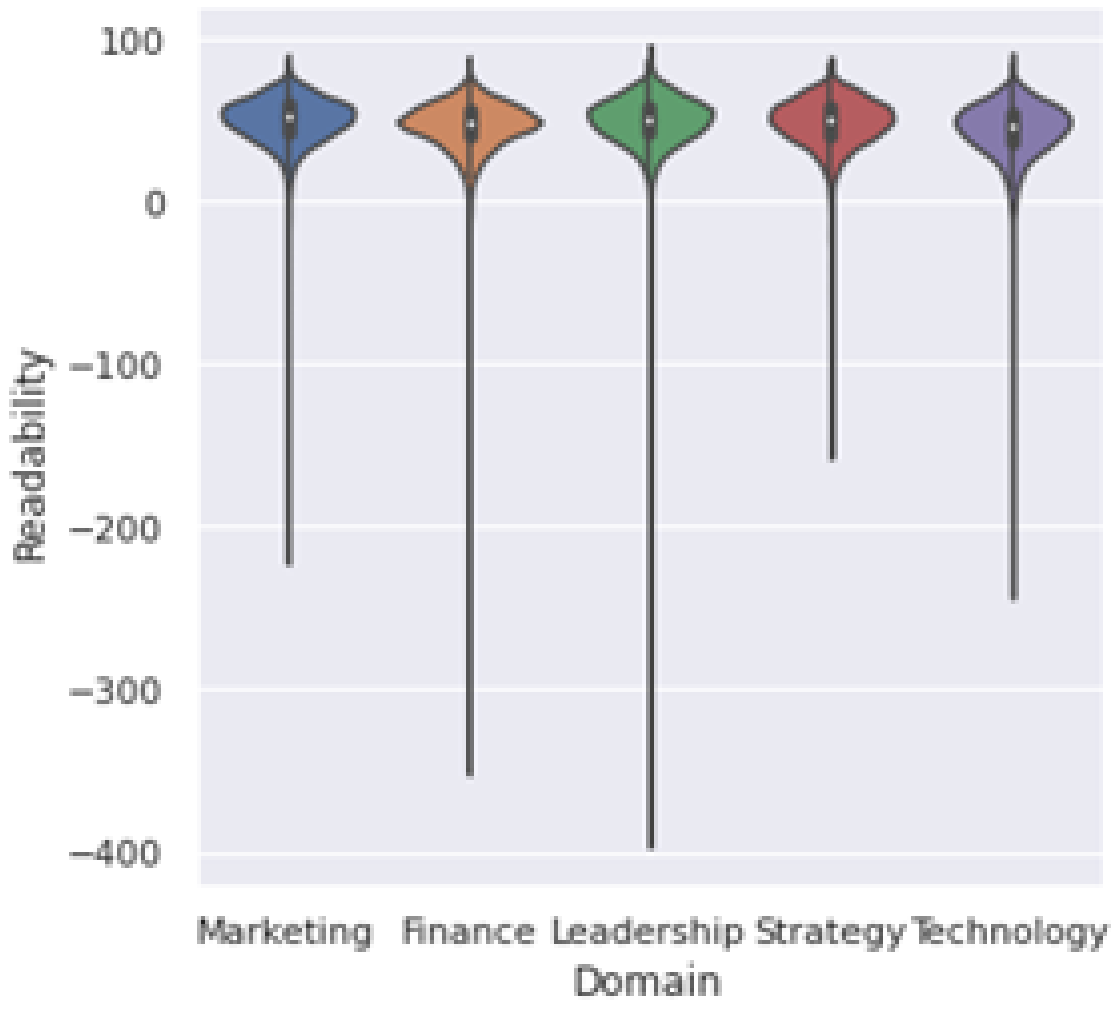

**Fig. 12.** Domain wise Readability distribution

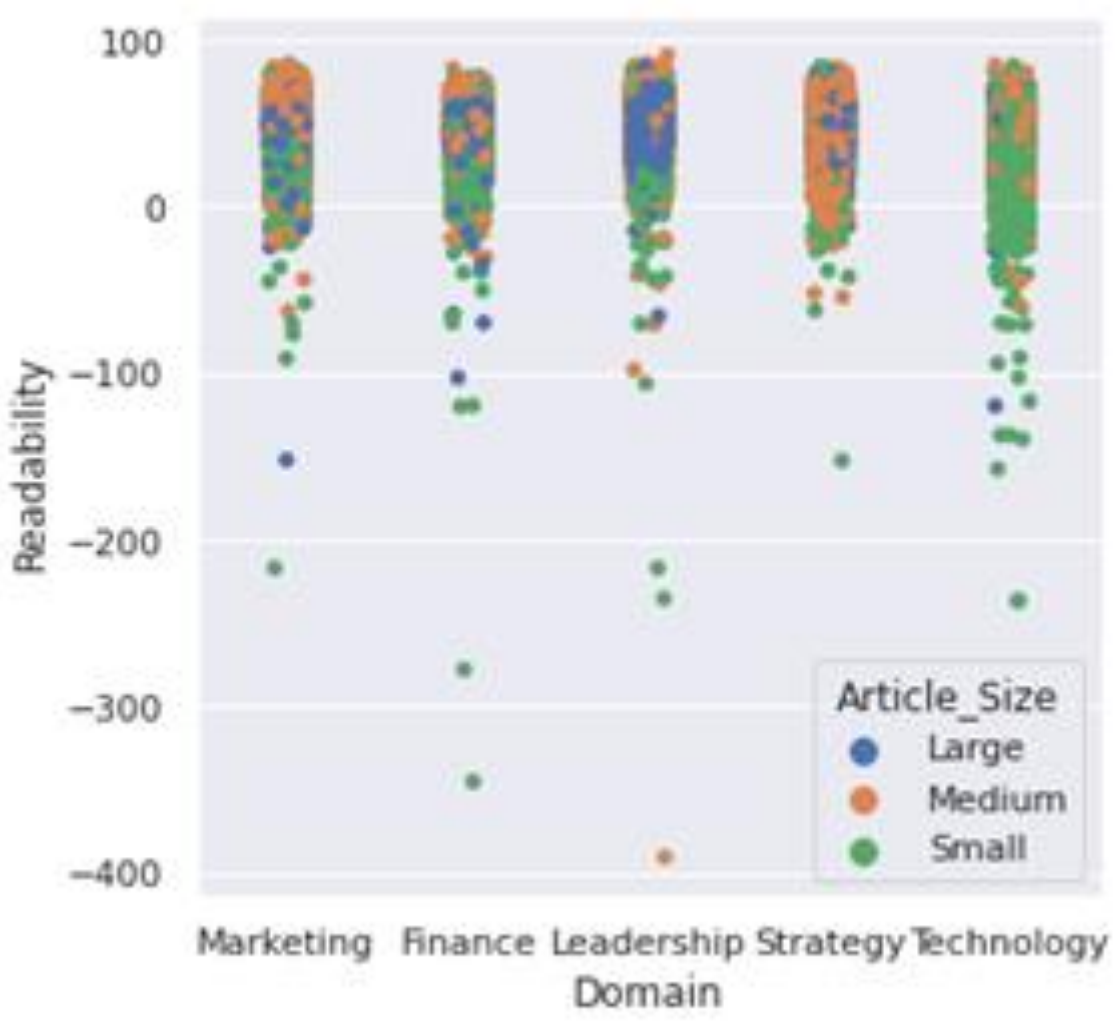

**Fig. 13.** Domain and Article wise Readability distribution

Most of the articles fall into College level reading ease followed by Grade 10. Also, we have almost zero percentage of Grade 5 level.

**Table 5.** Domain wise Readability class distribution

| F_Readability | Profes-sional | Grad-uate | Col-lege | Gra de 10 | Gra de 8 | Gra de 7 | Gra de 6 | Gra de 5 |
|---|---|---|---|---|---|---|---|---|
| **Finance** | 97 | 447 | 2089 | 1312 | 572 | 82 | 1 | 0 |
| **Leadership** | 71 | 425 | 2495 | 1882 | 1075 | 291 | 27 | 1 |
| **Marketing** | 106 | 450 | 2928 | 2303 | 1341 | 325 | 14 | 0 |
| **Strategy** | 77 | 432 | 2124 | 1491 | 888 | 183 | 10 | 0 |
| **Technology** | 241 | 781 | 2738 | 1517 | 706 | 104 | 5 | 0 |

### 5.4 Difficult-Word Density

Out of total words of the article, words with two or more syllables were defined as difficult words. Accordingly, four classes were created, viz., Basic, Elementary, Intermediate, and Advanced.

## 6 Recommendation of Most Suitable Authors

The articles were categorized in different classes of Readability, Domain, MBTI and Difficult-Word Density. Thereafter the different classes were encoded using OHE. Similarly, the sample text was categorized using OHE.

```
recommendations('Predicted Author')

[4761, 9570, 4760, 4759]

['Mark Abell', 'Roberta Holland', 'Eyal Shinar', 'Sona Jepsen']
```

**Fig. 14.** Recommendations model output

```
X1[(X1.shape[0]-1):X1.shape[0]]
```

| Cleaned_Author | ESFP | INFJ | INFP | INTJ | INTP | College | Grade_10 | Grade_5 | Grade_6 | Grade_7 | Grade_8 | Graduate | Professional | Finance | Leadership | Marketing | Strategy | Technology | Advanced | Basic | Elementary | Intermediate |
|---|---|---|---|---|---|---|---|---|---|---|---|---|---|---|---|---|---|---|---|---|---|---|
| **Predicted Author** | 0 | 0 | 0 | 1 | 0 | 1 | 0 | 0 | 0 | 0 | 0 | 0 | 0 | 1 | 0 | 0 | 0 | 0 | 0 | 0 | 1 | 0 |

```
X1[4761:4762]
```

| Cleaned_Author | ESFP | INFJ | INFP | INTJ | INTP | College | Grade_10 | Grade_5 | Grade_6 | Grade_7 | Grade_8 | Graduate | Professional | Finance | Leadership | Marketing | Strategy | Technology | Advanced | Basic | Elementary | Intermediate |
|---|---|---|---|---|---|---|---|---|---|---|---|---|---|---|---|---|---|---|---|---|---|---|
| **Mark Abell** | 0 | 0 | 0 | 1 | 0 | 1 | 0 | 0 | 0 | 0 | 0 | 0 | 0 | 1 | 0 | 0 | 0 | 0 | 0 | 0 | 1 | 0 |

**Fig. 15.** Recommendations model output comparison with input features

The Similarity Score using Cosine similarity was calculated for OHE matrices of corpora and the sample text. Based on the maximum cosine similarity value, the author(s) written the most similar text as compared with the sample text were recommended.

## 7 Business Impact

We believe that our solution is a good attempt to achieve the intended business outcome. The business content writing space is a fast evolving and dynamic field that is attracting multiple authors and content professionals - which will, in turn, only complicate the problem of identifying the best writers for a given business requirement. The use of AI and ML techniques can bring scale and relevance, while improving speed, accuracy, and depth and eliminate bias due to human errors and limitations.

# 8 Recommendations

We strongly believe that this solution has the potential to eliminate manual project allocation processes. Content creators will be able to create their profiles on various creator platforms and upload their entire portfolios for machine-based evaluation and cataloging without worrying about breach of intellectual property or infringement of creative license.

We recommend using (a further refined version of) our solution and building a platform that can be leveraged by all types of businesses including creative agencies, consulting companies, content platforms, freelancing platforms, etc. This will open up multiple opportunities by bridging the gap between content creators and content seekers and democratize the content space that is currently run by closed groups with major barriers to entry. It can help companies drive down content creation costs and while dramatically shooting up content recency and relevancy.

# 9 Limitations of the Solution

Just like any other machine learning solution, our solution is only as good as the data we have used to train it. For this project, we have sourced data from only two portals that usually do not marry branded content. This means our data could be biased (e.g., based on the editorial guidelines and the demographic target audience that those portals cater to), and not a true representation of the different types of writing styles that clients in the real-world might be seeking.

Our solution has been trained to predict writers for just a handful of business domains detailed above. While this was a conscious decision for the scope of this project, this is also a limiting factor as our model has not been exposed to diverse data. This can lead to biased predictions. Moreover, as described above, while our training data is multi-faceted, it is also imbalanced.

Modern-day business content writing requirements are not limited to long-form articles and blogs (like the ones we have used for our project). Content marketing strategies demand various other forms of content writing (and copywriting) like social media posts, infographics, research reports, whitepapers, technical case studies, PoVs, business plans, video scripts, and more. At this stage, our solution is not capable of identifying writers for these varied content formats.

Our solution also assumes that clients (content seekers) will have a sample write-up for the kind of output they are seeking; and this will be considered as the input for our production. However, this might not be true in the real world. We understand that clients rarely have reference write-ups. Instead, they have creative briefs and RFPs; or they simply prefer to talk to a writer and share a brief with him or her. They also prefer to have a conversation to explain their requirements.

## 10 Ideas to Enhance the Solution

The solution can be improved in multiple ways. Some of these could have a significant amount of improvement, while some others are simple enhancements that can have an incremental effect in the accuracy, applicability, and relevance. Here are the top four from our wish list:

1. **More Data:** While no amount of data can be termed as "enough data", we believe our solution can be far more relevant and real-world-ready with diverse data with more classes, writing styles and formats. Data from various sources across geography, demography, domains (content and vocabulary for specific domain), and tone of voice should be used. Special focus on portals that carry "branded" or "sponsored" content would bring in a lot more variety and hence a lot more real-life learning for our models.
2. **Different Models:** Use of different pretrained embeddings and transfer learning from pretrained models can be further explored to improve the domain classification results. This can include various BERT based pretrained models, DistilBERT, ALBERTA, etc. Fine Tuning pretrained transformer-based text classification models from HuggingFace 🤗 can also help improve the domain classification models.
3. **More Parameters:** The quest for finding the "best writer" is not an easy one. This is because there is no real definition for good writing. While this project uses some of the most basic parameters, a real-world writer needs to be examined on various other parameters. In the world of digital content, search engine optimization, and action-oriented writing, a good writer is expected to have mastered a lot more parameters like Grammar, Subject Matter, Formatting, Consistency, Structure, Call to Action, Product Placement, Depth and Recency of Research, Redundancy, Hyperlinks (and interlinks), use of Search Keywords, SEO Best Practices, Originality (plagiarism-free), and Adherence to the brand's Tone of Voice and Style Guide.
4. **More Computation Power:** Our models use Deep learning consisting of neural networks with multiple hidden layers. In addition, we are also using unsupervised learning algorithms for obtaining vector representations for words. These correspond to particularly demanding needs in terms of computational resources. As we plan to use more and more data, with a higher number of features and parameters, we believe that we will be requiring high-performance computational resources that can support multiple epochs and the corresponding long training times.

## 11 Key Learnings from the Process

The process gave us a great opportunity to plan and implement the entire project lifecycle - from problem identification and data collation to implementation and execution. As students, it was also an ideal setting to identify our interests, build on our strengths and work on our weaknesses.

The team has been learning and implementing new ideas and exploring different approaches throughout the project duration. We have also realized that several machine learning libraries, DL methods, NLP approaches, are dynamic and one needs to learn continuously. We learned that knowledge is the key to success and good data is divine.

*Note: The project was fully conceptualized, coded, and delivered remotely, and in the backdrop of a pandemic, it allowed us to develop useful skills for remote collaboration, communication, and teamwork.*

## 12 Ideas for Future Projects

The confluence of machine learning and human languages is an interesting and extremely dynamic space. In order to stay relevant, we will have to stay abreast with the developments in the field and keep innovating. For our next project, we believe we will have a lot more maturity. We would like to spend more time with data identification and data collation.